# Evaluating Vision Language Models (VLMs) for Radiology: A Comprehensive Analysis

Frank Li[1], Hari Trivedi[1], Bardia Khosravi[2], Theo Dapamede[1], Mohammadreza Chavoshi[1], Abdulhameed Dere[3], Rohan Satya Isaac[1], Aawez Mansuri[1], Janice Newsome[1], Saptarshi Purkayastha[4], Judy Gichoya[1]

[1]Department of Radiology, Emory University, Atlanta, GA, USA

[2]Department of Radiology, Mayo Clinic, Rochester, MN, USA

[3]Faculty of Clinical Sciences, College of Health Sciences, University of Ilorin, Ilorin, Nigeria

[4]Luddy School of Informatics, Computing, and Engineering, Indiana University Indianapolis, Indianapolis, IN, USA

**Abstract**

Foundation models—trained on vast amounts of data using self-supervised techniques—have emerged as a promising frontier for advancing artificial intelligence (AI) applications in medicine. This study evaluates three different vision-language foundation models (RAD-DINO, CheXagent, and BiomedCLIP) on their ability to capture fine-grained imaging features for radiology tasks. The models were assessed across classification, segmentation, and regression tasks for pneumothorax and cardiomegaly on chest radiographs. Self-supervised RAD-DINO consistently excelled in segmentation tasks, while text-supervised CheXagent demonstrated superior classification performance. BiomedCLIP showed inconsistent performance across tasks. A custom segmentation model that integrates global and local features substantially improved performance for all foundation models, particularly for challenging pneumothorax segmentation. The findings highlight that pre-training methodology significantly influences model performance on specific downstream tasks. For fine-grained segmentation tasks, models trained without text supervision performed better, while text-supervised models offered advantages in classification and interpretability. These insights provide guidance for selecting foundation models based on specific clinical applications in radiology.

1. **Introduction**

Recent years have witnessed the rapid development of deep learning models for medical imaging facilitated by availability of computational resources and large annotated datasets. Many medical applications developed using supervised learning approaches have shown high-level performance, sometimes surpassing that of human experts. Despite their usefulness, these training paradigms exhibit several limitations. These include limited explainability of black box models, the burden of annotating large datasets which are also characterized by extreme class imbalance, focus on narrow predictive tasks, and limited ability to handle multimodal data. Additionally, these models demonstrate persistent performance disparities (underdiagnosis and overdiagnosis) across several demographic and clinical subgroups (1). Foundation models, trained on vast amounts of data using self-supervised techniques, represent the next frontier for artificial intelligence (AI) in medicine aiming to address and overcome the above limitations (2,3). Foundation models are characterized by their ability to perform tasks they weren't specifically trained for and effectively handle new information without requiring retraining or with minimal fine-tuning using limited data. In addition, they can be adapted to a wide range of downstream tasks (4).

Vision-language models (VLMs) have emerged as a particularly promising category of foundation models that seamlessly integrate both visual and textual understanding. Recent VLMs are typically structured around two essential components: the vision encoder and the text decoder. The vision encoder extracts meaningful features from images, while the text decoder processes these extracted features, combined with user prompts to provide answers in human comprehensible language. This sophisticated dual-modality capability enables intuitive natural language interaction, allowing users to communicate with these systems through ordinary text instructions. This versatility of VLMs makes them invaluable across diverse medical applications, including disease diagnosis, interactive interpretation of medical images, generation of clinical reports, and phrase grounding (precisely locating specific anatomical or pathological features within images) (5). Additionally, VLMs excel at extracting imaging features for various downstream tasks, leveraging their broad knowledge acquired during pre-training. Models built on these extracted features have been shown to have superior performance with reduced training data requirements, thus improving efficiency and reducing the burden of intensive human labeling (6–10).

Current medical VLMs generally develop their vision encoders through one of two pre-training approaches. The first method employs text-supervised pre-training, utilizing contrastive learning techniques (e.g. CLIP (11)) to align text and image features within a shared latent space (8,12–16). The second approach concentrates exclusively on visual data through techniques such as masked image modeling and self-supervised instance discrimination, enabling the model to develop robust visual representations independent of textual guidance (6,7,9,10). The lack of a standardized end to end development approach has resulted in available VLMs being trained with different dataset types and sizes (6,16). This inconsistency complicates the process of comparing performance across different tasks and determining which VLM is most suitable for a specific task.

Despite their promising capabilities, VLMs face several important limitations. The vision encoders of most models learn to understand images through contrastive learning, which helps them associate images with corresponding text descriptions. However, this learning process can be significantly limited when text descriptions lack sufficient detail as may occur when the model is trained on the abbreviated "Impression" section of a radiology report versus the more detailed "Findings" section. This can lead to representation collapse (17), when radiology reports fail to document all important findings. In this case the model may oversimplify its understanding of images to match them with text, potentially missing crucial differences between similar cases. Without understanding intra-class variations, these image encoders may not generalize to broader healthcare applications, eventually requiring new training for each specific application.

This paper addresses an existing knowledge gap in determining the suitable VLM for specific radiology tasks, considering known limitations of VLMs and their variability in training data type and scale. We evaluate the ability of VLMs to capture fine-grained, disease-related imaging features – necessary for robust applications in radiology, and their impact on the downstream performance for classification and segmentation tasks. Segmentation tasks provide a rigorous assessment of a model's visual discrimination abilities at a fine-grained level and have not been previously described in published VLMs studies.

## 2. Methods

### 2.1. Experimental setup

Our objective was to evaluate performance of the image encoders of foundation models on two segmentation tasks - pneumothorax and cardiomegaly. For complementary evaluation metrics, we performed classification and regression to establish baseline performance measures of the models' global understanding of these pathologies.

To accomplish this, we extracted embeddings from three representative foundation models (**Table 1**) that utilize vision transformers (ViT) as their image encoders and use CXRs as a component of their training data. RAD-DINO (6) operates without text supervision; CheXagent (12) employs text supervision through contrastive learning; and BiomedCLIP (18), incorporates both multimodal and text supervision approaches.

**Table 1.** Summary of the three foundation models evaluated in this study.

| Model | Year | Multi-imaging Modal | Image Encoder | Pretraining Method | Text Supervised | Sample Size | Input Image Size | Embedding Size Global ([CLS]) | Embedding Size Patch (C*H*W) |
|---|---|---|---|---|---|---|---|---|---|
| **RAD-DINO** | 2024 | No | ViT-B/14 | MIM SSID | No | 838,336 | 518*518 | 768 | 768*37*37 |
| **CheXagent** | 2024 | No | SigLIP-Large | CLIP | Yes | 1,077,494 | 512*512 | 1024 | 1024*32*32 |
| **BiomedCLIP** | 2023 | Yes | ViT-B/16 | CLIP | Yes | 15,000,000 | 224*224 | 512 | 768*14*14 |

* MIM: Masked Image Modeling; SSID: Self-Supervised Instance Discrimination; CLIP: Contrastive Language-Image Pre-Training; C: Number of channels; H: Height of the image; W: Width of the image.

As a first step, we extracted both patch embeddings and global ([[CLS]]) embeddings from the three selected VLMs. Patch embeddings represent detailed local feature representations, while [CLS] embeddings capture the global representation of images. It's important to note that the extracted [CLS] embeddings constitute the final outputs of the vision encoders.

For pneumothorax assessment, we conducted three analyses:

1. Segmentation using linear probing (19) on the patch embeddings.

2. Segmentation using a custom model that integrates patch and [CLS] embeddings via cross-attention.

3. Classification using linear probing on the [CLS] embeddings (19)

For cardiomegaly evaluation, we performed four distinct analyses:

1. Heart segmentation using linear probing on patch embeddings.

2. Heart segmentation using a custom model integrating patch and [CLS] embeddings via cross-attention.

3. Cardiomegaly classification using linear probing on [CLS] embeddings (19)

4. Cardiothoracic ratio (CTR) regression using linear probing on [CLS] embeddings.

**Figure 1** illustrates the experimental setup and the architecture of our custom segmentation model. Implementation details of all models are provided in the **Supplementary Material**. To assess the foundation models' ability to encode essential information for downstream tasks, we employed simpler approaches such as linear probing and our custom segmentator rather than more complex architectures (e.g. U-Net).

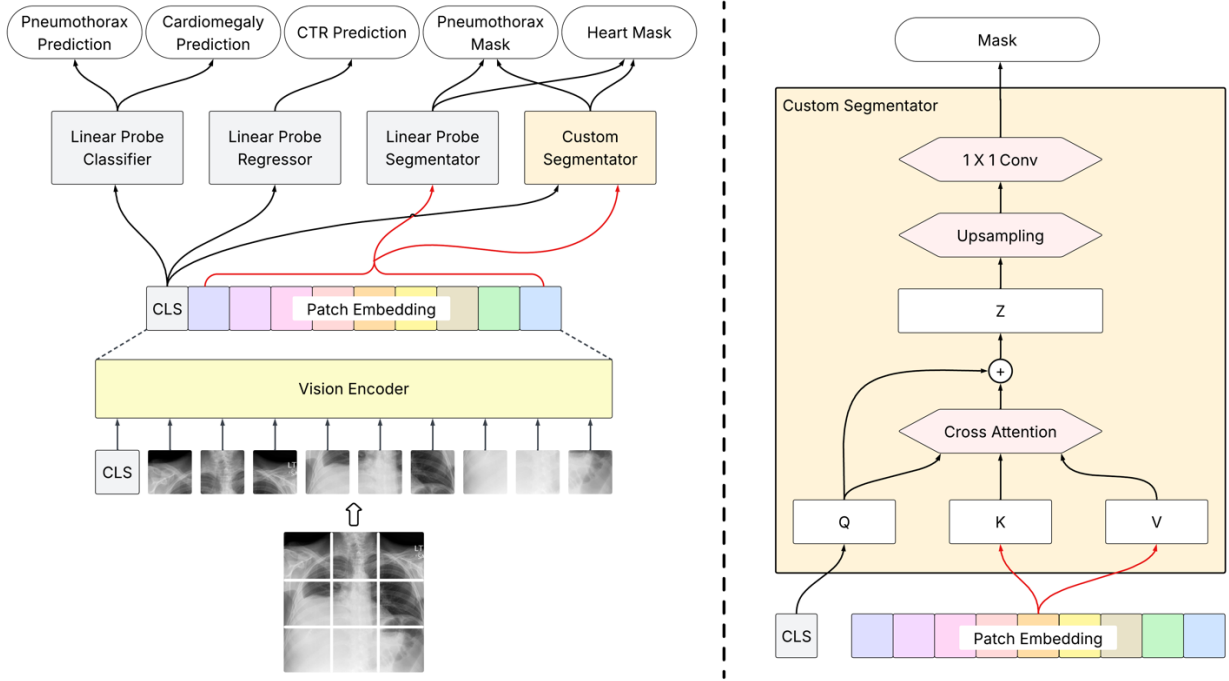

**Figure 1.** Experimental setup (left) illustrating the key components of our study methodology, alongside the architecture diagram of our custom segmentation model (right).

2.2. Data Preparation

For pneumothorax evaluations, we utilized the SIIM-ACR Pneumothorax dataset (20), which comprises 10,675 CXRs. For segmentation tasks, the models were trained exclusively on positive cases (n=2,379). For the classification task, we down-sampled negative cases to create a balanced dataset (n=4,758). We balanced the data to eliminate prevalence effects (21).

For cardiomegaly evaluation, we sampled a cohort of 30,000 posterior-anterior (PA) view CXRs from a private dataset collected at Emory Healthcare. This balanced cohort contains an equal distribution of positive and negative cases (50% each) based on the report labels. As with the pneumothorax dataset, we ensured balanced classification data to prevent prevalence bias in our evaluations. Heart and lung masks were derived following the methodologies established in CheXmask (18). We calculated the cardiothoracic ratio (CTR) by dividing the projection of the heart mask on the X-axis by the combined projections of heart and lung masks on the same axis (**Equation 1**).

We implemented an 72-8-20 split at the level of the patient to separate training, validation, and testing datasets for both pneumothorax and cardiomegaly evaluations.

**Equation 1**

$$CTR = Projection_{Heart}/Projection_{Heart+Lung}$$

## 2.3. Training Setup

The training setup for all downstream models utilized a batch size of 16 and a learning rate of 1e-4, running on a Quadro RTX 6000 GPU. To prevent overfitting, we implemented early stopping to terminate training after 10 epochs of minimal improvement (delta threshold of 0.0001). The implementation leveraged PyTorch and PyTorch Lightning frameworks for efficient model development and training.

## 2.4. Evaluation of Downstream Models

Evaluation metrics were tailored to each model type. For segmentation models, Dice coefficient and Intersection over Union (IoU) were employed to assess spatial overlap accuracy. Classification models were evaluated using Area Under the Receiver Operating Characteristic curve (AUROC) and Area Under the Precision-Recall curve (AUPRC), which comprehensively measure discrimination ability across thresholds. Regression models were assessed using coefficient of determination ($R^2$) and MSE to quantify prediction accuracy and deviation. To establish statistical reliability, 10,000 bootstrap resamples of the test data were performed to construct 95% confidence intervals for all performance metrics. Statistical significance was determined using a two-sample bootstrapping test with a significance threshold of $p < 0.0167$ with Bonferroni correction.

## 3. Results

We evaluated three foundation models—CheXagent, RAD-DINO, and BiomedCLIP—across four medical imaging tasks: segmentation using linear probing, segmentation using custom models, classification, and regression. The results of the evaluations are summarized in **Table 2** and **Figure 2** and the results of pairwise comparisons between the foundation models are documented in **Supplementary Material/Table S1**.

**Table 2.** Performance comparison of foundation models (CheXagent, RAD-DINO, and BiomedCLIP) on medical imaging tasks. Performance metrics are presented with 95% confidence intervals shown in parentheses.

| Task | Scenario | Metric | CheXagent | RAD-DINO | BiomedCLIP |
|---|---|---|---|---|---|
| **Segmentation (Linear Probing)** | Pneumothorax | Dice | 0.293 (0.270 - 0.317) | 0.257 (0.233 - 0.281) | 0.084 (0.069 - 0.099) |
| | | IoU | 0.204 (0.185 - 0.222) | 0.179 (0.161 - 0.198) | 0.054 (0.044 - 0.064) |
| | Heart | Dice | 0.906 (0.905 - 0.907) | 0.945 (0.944 - 0.946) | 0.917 (0.916 - 0.918) |
| | | IoU | 0.830 (0.828 - 0.831) | 0.897 (0.896 - 0.898) | 0.848 (0.847 - 0.850) |
| **Segmentation (Custom Model)** | Pneumothorax | Dice | 0.364 (0.337 - 0.391) | 0.424 (0.398 - 0.449) | 0.250 (0.227 - 0.273) |
| | | IoU | 0.268 (0.245 - 0.290) | 0.312 (0.290 - 0.334) | 0.171 (0.154 - 0.189) |
| | Heart | Dice | 0.955 (0.954 - 0.955) | 0.962 (0.962 - 0.963) | 0.956 (0.956 - 0.957) |
| | | IoU | 0.914 (0.913 - 0.915) | 0.928 (0.927 - 0.929) | 0.917 (0.916 - 0.918) |
| **Classification** | Pneumothorax | AUROC | 0.955 (0.943 - 0.967) | 0.916 (0.898 - 0.934) | 0.860 (0.836 - 0.882) |
| | | AUPRC | 0.955 (0.940 - 0.968) | 0.905 (0.876 - 0.931) | 0.859 (0.828 - 0.887) |
| | Cardiomegaly | AUROC | 0.941 (0.935 - 0.947) | 0.928 (0.921 - 0.934) | 0.907 (0.899 - 0.914) |
| | | AUPRC | 0.932 (0.924 - 0.941) | 0.919 (0.910 - 0.928) | 0.894 (0.882 - 0.905) |
| **Regression** | CTR | $R^2$ | 0.825 (0.817 - 0.833) | 0.830 (0.821 - 0.839) | 0.649 (0.633 - 0.664) |
| | | MSE | 0.001 (0.001 - 0.001) | 0.001 (0.001 - 0.001) | 0.02 (.002 - 0.002) |

### 3.1. Segmentation (Linear Probing)

In the linear probing segmentation task for pneumothorax, CheXagent outperformed other models with a Dice coefficient of 0.293 (95% CI: 0.270-0.317) and IoU of 0.204 (95% CI: 0.185-0.222). RAD-DINO ranked second (Dice: 0.257, IoU: 0.179), while BiomedCLIP showed significantly lower performance (Dice: 0.084, IoU: 0.054). The performance differences between CheXagent and BiomedCLIP, as well as between RAD-DINO and BiomedCLIP, were statistically significant (p < 0.0001). However, the difference between CheXagent and RAD-DINO did not reach statistical significance for either Dice (p = 0.039) or IoU (p = 0.067).

For heart segmentation, RAD-DINO demonstrated superior performance, achieving the highest Dice coefficient (0.945, 95% CI: 0.944-0.946) and IoU (0.897, 95% CI: 0.896-0.898). CheXagent exhibited the lowest performance (Dice: 0.906, IoU: 0.830), while BiomedCLIP achieved intermediate results (Dice: 0.917, IoU: 0.848). The differences between all model pairs were statistically significant (p < 0.0001).

### 3.2. Segmentation (Custom Model)

For pneumothorax segmentation with custom models, RAD-DINO achieved the best performance (Dice: 0.424, 95% CI: 0.398-0.449; IoU: 0.312, 95% CI: 0.290-0.334), followed by CheXagent

(Dice: 0.364, IoU: 0.268) and BiomedCLIP (Dice: 0.250, IoU: 0.171). All pairwise differences were statistically significant, with p = 0.0014 for the CheXagent vs. RAD-DINO Dice comparison, p = 0.0056 for the IoU comparison, and p < 0.0001 for all other comparisons.

When using custom models for heart segmentation, RAD-DINO again demonstrated superior performance with the highest Dice coefficient (0.962, 95% CI: 0.962-0.963) and IoU (0.928, 95% CI: 0.927-0.929). BiomedCLIP showed slightly better performance than CheXagent, with Dice coefficients of 0.956 vs. 0.955 and IoU of 0.917 vs. 0.914, respectively. All pairwise differences were statistically significant (p < 0.0001), though the absolute differences were small.

### 3.3. Linear Probing vs. Custom Models for Segmentation Tasks

The transition from linear probing to custom model approaches for segmentation demonstrates substantial performance improvements across all foundation models. For pneumothorax segmentation (**Figure 3a**), RAD-DINO showed the most dramatic improvement, with Dice scores increasing from 0.257 to 0.424 (a 65.0% relative increase) and IoU from 0.179 to 0.312 (a 74.3% improvement). Similarly, CheXagent's performance improved from 0.293 to 0.364 Dice (24% increase) and 0.204 to 0.268 IoU (31.4% increase), while BiomedCLIP exhibited the largest relative gains, with Dice increasing from 0.084 to 0.250 (197.6% increase) and IoU from 0.054 to 0.171 (216.7% increase).

For heart segmentation (**Figure 3b**), RAD-DINO maintained its superior performance while improving from 0.945 to 0.962 Dice (1.8% increase) and 0.897 to 0.928 IoU (3.5% increase). CheXagent showed more substantial gains in heart segmentation, with Dice increasing from 0.906 to 0.955 (5.4% increase) and IoU from 0.830 to 0.914 (10.1% increase), while BiomedCLIP improved from 0.917 to 0.956 Dice (4.3% increase) and 0.848 to 0.917 IoU (8.1% increase).

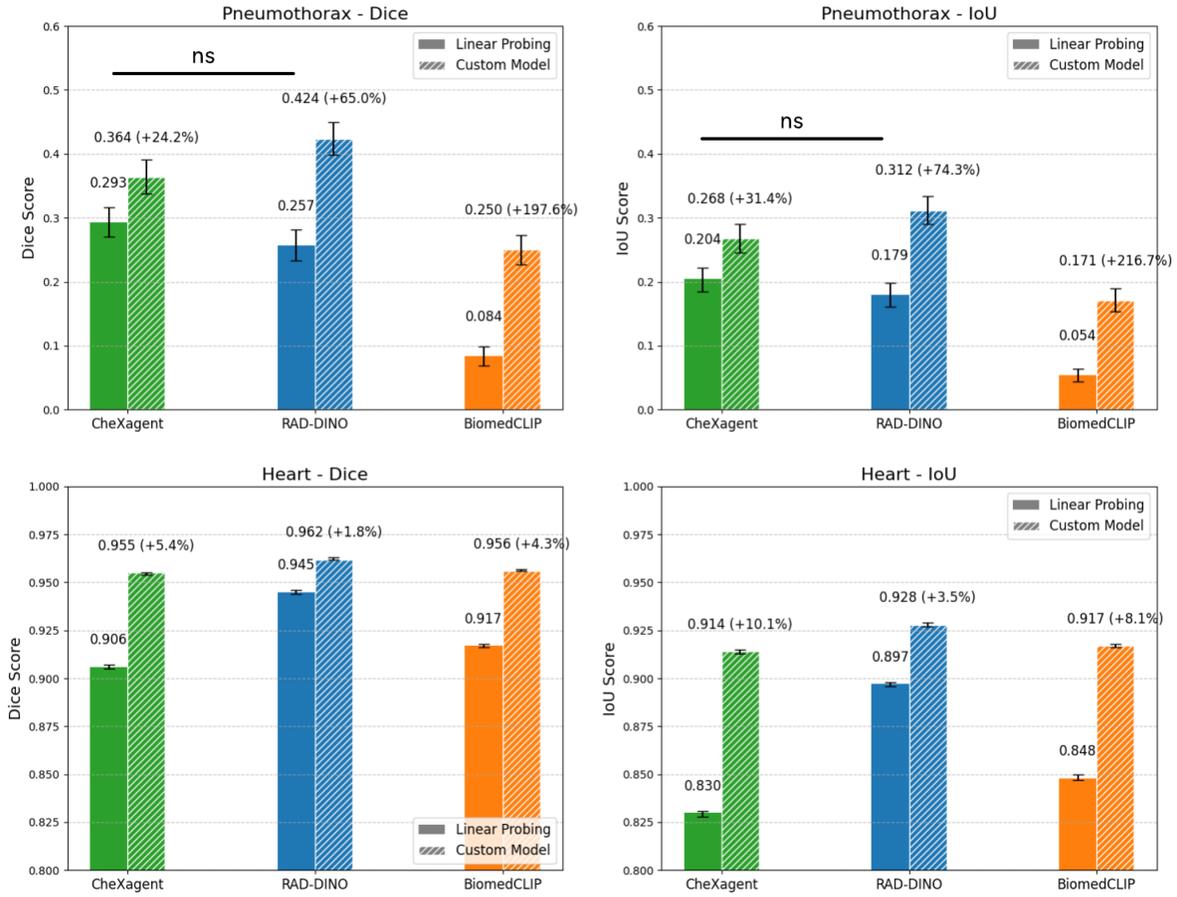

**Figure 2.** Performance comparison of segmentation approaches on chest X-ray images. Pneumothorax (top) and heart (bottom) segmentation performance measured by Dice coefficient (left) and IoU (right) using linear probing (solid bars) versus custom models (hatched bars). Percentage improvements with custom models are shown above each pair. The results are shown with 95% confidence interval error bars. Non-significant differences are marked "ns".

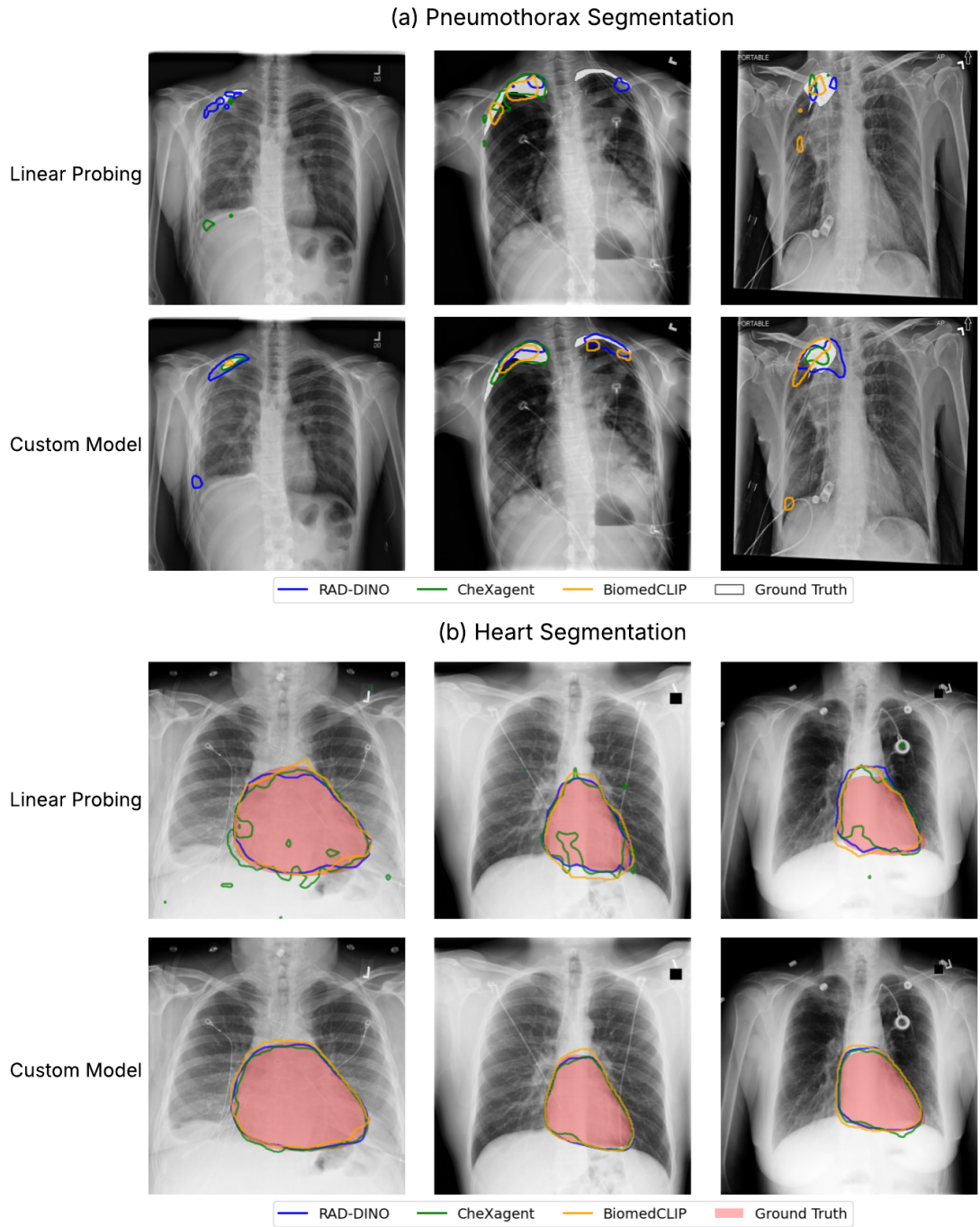

**Figure 3.** Comparative performance visualization of linear probing versus custom segmentation techniques across various foundation models for medical image segmentation tasks: (a) pneumothorax segmentation and (b) heart segmentation.

### 3.4. Classification

For pneumothorax classification, CheXagent achieved the highest performance (AUPRC: 0.955, 95% CI: 0.940-0.968; AUROC: 0.955, 95% CI: 0.943-0.967), followed by RAD-DINO (AUPRC: 0.905, AUROC: 0.916) and BiomedCLIP (AUPRC: 0.859, AUROC: 0.860). The differences between CheXagent and the other models were statistically significant for both metrics ($p < 0.0001$). The difference between RAD-DINO and BiomedCLIP was significant for AUROC ($p = 0.0002$) but not for AUPRC ($p = 0.027$).

In the cardiomegaly classification task, CheXagent demonstrated the highest performance with an AUPRC of 0.932 (95% CI: 0.924-0.941) and AUROC of 0.941 (95% CI: 0.935-0.947). RAD-DINO showed intermediate performance (AUPRC: 0.919, AUROC: 0.928), while BiomedCLIP had the lowest scores (AUPRC: 0.894, AUROC: 0.907). The difference between CheXagent and RAD-DINO was not statistically significant for AUPRC ($p = 0.0382$) but was significant for AUROC ($p = 0.001$). All other pairwise differences were statistically significant ($p < 0.0001$).

### 3.5. Regression

In the cardiothoracic ratio (CTR) regression task, RAD-DINO achieved the highest $R^2$ value (0.830, 95% CI: 0.821-0.839), slightly outperforming CheXagent (0.825, 95% CI: 0.817-0.833), though this difference was not statistically significant ($p = 0.4314$). Both models significantly outperformed BiomedCLIP ($R^2$: 0.649, 95% CI: 0.633-0.664) with $p < 0.0001$.

For MSE in CTR regression, both CheXagent and RAD-DINO achieved a value of 0.001 (with identical confidence intervals), showing no statistically significant difference ($p = 0.3622$). BiomedCLIP demonstrated significantly higher error (MSE: 0.002) compared to both other models ($p < 0.0001$).

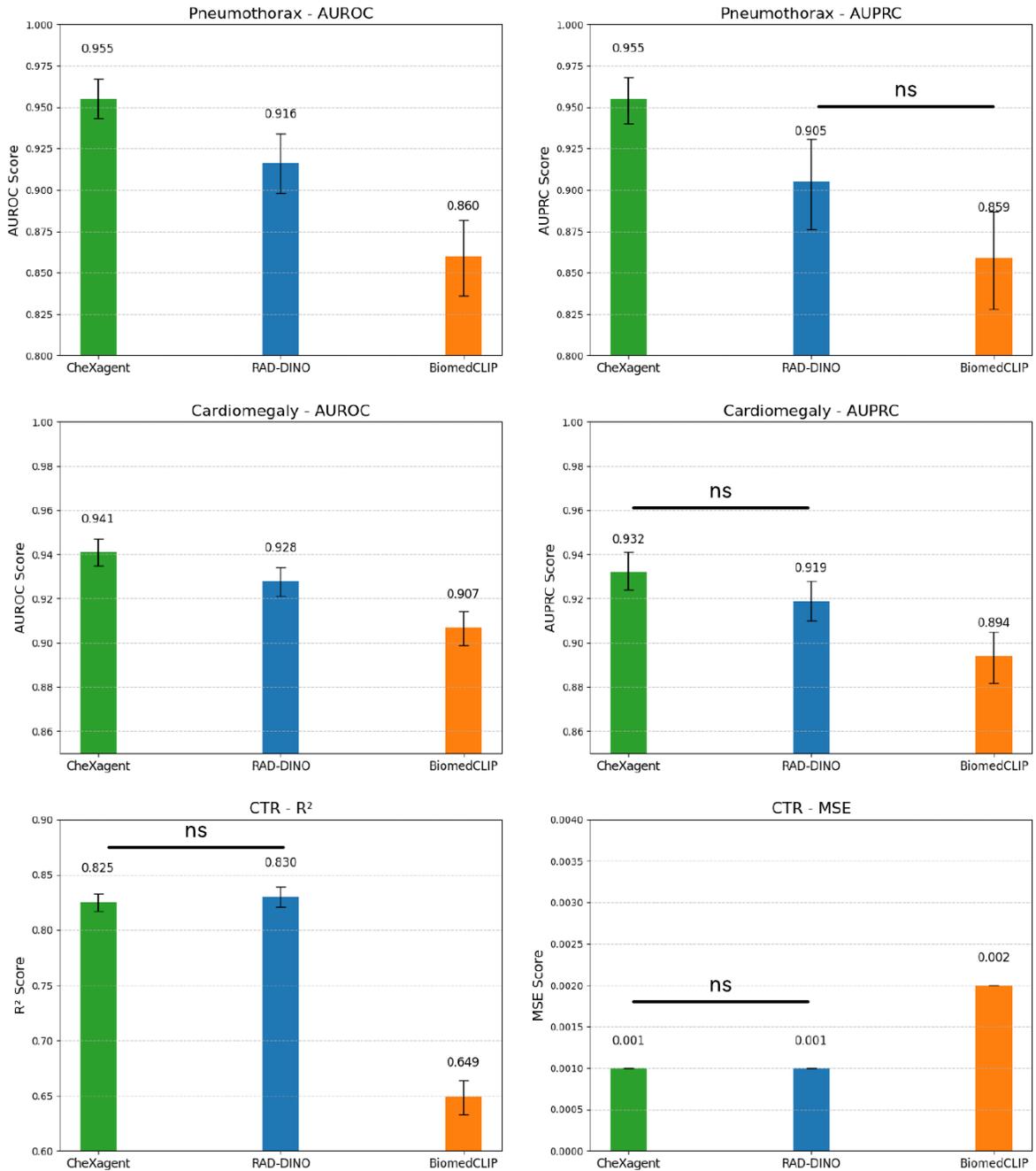

**Figure 4.** Performance comparison of classification and regression approaches on chest X-ray images. Pneumothorax classification (top row) performance measured by AUROC (left) and AUPRC (right). Cardiomegaly classification (middle row) performance measured by AUROC (left) and AUPRC (right). CTR regression (bottom row) performance measured by $R^2$ (left) and

MSE (right). The results are shown with 95% confidence interval error bars. Non-significant differences are marked "ns".

## 4. Discussion

Our evaluation of three foundation models (CheXagent, RAD-DINO, and BiomedCLIP) across two medical imaging tasks reveals important patterns in their performance and capabilities. All models achieved good segmentation performance despite a small sample size of the pneumothorax dataset and the use of lightweight models (e.g. linear probing) rather than full fine-tuning. This finding supports the potential for broad application of foundation models for variable medical imaging tasks (even when the model was not trained for that specific task) with small datasets while remaining computationally efficient.

RAD-DINO, an image encoder only model provided consistently good performance across different downstream tasks. The pneumothorax segmentation results demonstrated a substantial difference between linear probing and the custom segmentation model performance. The significant improvement seen with the custom model for pneumothorax segmentation (65.0% for RAD-DINO, 24.2% for CheXagent, and 197.6% for BiomedCLIP) suggests that integrating global context information with local features greatly enhances segmentation performance, particularly for BiomedCLIP. While CheXagent demonstrated marginally better performance in linear probing segmentation for pneumothorax (though not statistically significant compared to RAD-DINO), RAD-DINO consistently outperformed all competitors when implementing the custom segmentation model. This indicates that RAD-DINO's patch embeddings may contain richer local information that benefits from global context integration, despite not being explicitly trained with text supervision. Without text supervision, RAD-DINO was able to learn fine-grained imaging features which capture the essential information for downstream predictions. In contrast, text-supervised models like CheXagent excel in classification but show variable performance in segmentation tasks, suggesting potential representation collapse in certain domains.

We observed substantially higher performance across all three models for cardiomegaly segmentation compared to pneumothorax segmentation. This difference suggests that heart segmentation may be an inherently easier task than pneumothorax segmentation, possibly due to the heart's more consistent shape, size, and position within chest radiographs compared to the variable presentation of pneumothorax. The custom segmentation model showed more modest relative improvements for heart segmentation compared to pneumothorax segmentation. This smaller improvement differential could indicate that linear probing already captures most of the relevant features for heart segmentation, leaving less room for improvement through global context integration.

The classification results reveal that CheXagent, trained with text supervision through contrastive learning, demonstrated superior performance for both pneumothorax and cardiomegaly classification tasks. This suggests that text-supervised pre-training may produce global image representations ([CLS] embeddings) that effectively capture disease presence but may not preserve the fine-grained spatial information necessary for optimal segmentation. The CTR regression results showed RAD-DINO and CheXagent performing similarly well, while BiomedCLIP lagged significantly. This performance gap in regression tasks could indicate limitations in BiomedCLIP's ability to capture continuous variables or quantitative relationships.

BiomedCLIP had the worst comparative performance of the three models. It showed the most variable performance across tasks, performing relatively well in heart segmentation but poorly in pneumothorax segmentation and CTR regression. This inconsistency might stem from several factors: 1). BiomedCLIP may suffer from representation collapse, as evidenced by its worst performance among the three models on pneumothorax segmentation and CTR regression. 2). It's a multi-modal model trained on a full spectrum of medical images and not specialized in chest X-rays – implying that fine grained performance necessary for segmentation is obtained from models trained on disease specific datasets.  3). The training images, being published with academic articles, were likely selected from very representative images and potentially unable to capture the wide range of real-world datasets. 4. Its small input image size (224×224) during training may cause loss of important details as radiology images are highly pixelated.

Our findings suggest that the choice of pre-training approach significantly impacts model performance across different radiology tasks. RAD-DINO, which employs self-supervised pre-training without text supervision, consistently excelled in segmentation tasks. This suggests that self-supervised learning may develop more granular visual representations beneficial for detailed spatial understanding. CLIP-based models (like CheXagent and BiomedCLIP) learn more abstract, human-interpretable features, including cultural and semantic concepts that generalize across visual styles. In contrast, DINO-based models like RAD-DINO learn more granular, style-specific features (22). However, for medical imaging segmentation, abstract concepts appear less important than granular imaging features, which may explain RAD-DINO's consistent performance across tasks. The discrepancy between classification and segmentation performance highlights the importance of evaluating foundation models across diverse task types. It's worth noting that SIIM-ACR PTX was part of CheXagent's training data, which likely contributed to its strong performance in pneumothorax-related tasks compared to the cardiomegaly segmentation, but both had overall lower performance compared to RAD-DINO.

Additionally, we demonstrate that text-supervised foundation models benefit more from global [CLS] embedding integration with improvement in segmentation tasks. This is likely because these models were optimized with objectives calculated based on [CLS] embeddings during the pre-training stage. While text-supervised foundation models may be prone to representation collapse, they provide out-of-the-box ability for zero-shot inference and human interpretability. For broad radiology tasks, there is need for good performance on fine grained image tasks like segmentation combined with features like human interpretability which means both approaches are necessary.

Extracting embeddings for this project was challenging, highlighting other challenges of adopting this new training paradigm. Documentation of foundation models rarely specified image preprocessing requirements (like 16-bit or 8-bit PNG formats), and the pre-processors included with foundation models often can't handle variety of medical image formats—yet proper preprocessing is crucial for accurate results. Since users frequently need to evaluate the source code to figure out the necessary preprocessing steps and input format, this type of evaluation is nearly impossible with closed-source models. There is need to adapt or develop new standards to streamline this process. Additionally, since downstream tasks are typically limited to pre-defined options (e.g. Visual question answering (VQA), captioning, zero-shot classification, and segmentation), these restrict the flexibly of use for foundation models and requires in-depth understanding of model structures to extract specific outputs needed for different applications. For example, The BiomedCLIP paper illustrates this limitation—while successfully demonstrating classification capabilities through VQA, it fails to provide clear methodologies for extracting image embeddings from the image encoder for downstream segmentation tasks, leaving their potential performance unclear.

While our study provides valuable insights into foundation model capabilities, several limitations should be acknowledged. First, our evaluation focused specifically on chest radiographs and two clinical conditions. Future work should extend to other imaging modalities and pathologies to assess the generalizability of these findings. Second, the performance differences observed might be influenced by the specific architectures of the vision encoders beyond their pre-training approaches. Controlling for architectural variations could provide clearer insights into the impact of pre-training methodologies. Finally, our custom segmentation model, while demonstrating significant improvements, represents just one approach to integrating global and local features. Alternative integration methods might yield different performance patterns and should be explored in future research.

## 5. Conclusion

In conclusion, our results emphasize the importance of task-specific evaluation for foundation models in medical imaging. We demonstrate that the pre-training approach significantly influences the model's strengths and weaknesses across different downstream tasks, suggesting that hybrid approaches combining self-supervised and text-supervised learning might offer more comprehensive performance across diverse medical imaging applications.

Based on our findings, we offer the following recommendations:

1. If the goal is to extract fine-grained features – useful for tasks like segmentation, use foundation models pre-trained without text supervision (like RAD-DINO).

2. If the goal is to perform zero-shot inference, capture higher-level concepts, and have human interpretability, use foundation models pre-trained with text supervision (like CheXagent).

3. If foundation models pre-trained with text supervision are the only options and the goal is to perform tasks that require capturing fine-grained features (e.g., segmentation, regression), integrate the global ([CLS]) and patch embeddings as demonstrated in our custom segmentation model.

## 6. Declarations

### 6.1. Ethics Approval and Consent to Participate

This study was approved by the institutional review board of Emory University (STUDY00002276: Quality and Informatics Protocol) and informed consent was waived. The manuscript and the results were presented in a way that the patients cannot be identified.

### 6.2. Competing Interests

There is no conflict of interest for all authors.

### 6.3. Funding/Support

This study was supported by the AI Image Extraction Core, an Emory Integrated Core Facility. Dr. Gichoya is a 2022 Robert Wood Johnson Foundation Harold Amos Medical Faculty Development Program and declares support from Lacuna Fund (#67), Gordon and Betty Moore Foundation, NIH (NIBIB) MIDRC grant under contracts 75N92020C00008 and 75N92020C00021, NHLBI Award Number R01HL167811 and NIH common fund award 1R25OD039834-01.

### 6.4. Author's Contributions

All authors made a significant contribution to the work reported, whether that is in the conception, study design, execution, acquisition of data, analysis and interpretation, or in all these areas; took part in drafting, revising or critically reviewing the article; gave final approval of the version to be published; have agreed on the journal to which the article has been submitted; and agree to be accountable for all aspects of the work.

# Supplementary Material

## 1.1. Linear Probing for Segmentation

The linear probing segmentation model leverages patch embeddings from a pre-trained vision encoder. It features a single 1×1 convolutional layer that efficiently projects high-dimensional feature maps (e.g. 768 channels at 37×37 resolution for RAD-DINO) to a target class (e.g. pneumothorax), followed by bilinear upsampling to restore the original image resolution (e.g. 518×518 for RAD-DINO). This minimalist architecture helps determine whether rich semantic information already exists within the patch embeddings (1). The model functions as both a baseline and a reference point for ablation studies when evaluating more complex segmentation approaches, and utilizes binary cross-entropy with logits loss for training.

## 1.2. A Custom Segmentation Model

This custom model (**Figure 1**) enhances segmentation by combining local and global features from a pre-trained vision encoder. It processes patch embeddings through layer normalization, upsampling, and a two-layer CNN refiner, while simultaneously normalizing and projecting the CLS embedding separately. A cross-attention mechanism enables the CLS embedding to attend to the refined patch features, creating a global context that integrates with local features via a residual connection. This comprehensive representation is then upsampled to the target resolution and processed by a 1×1 convolutional layer to generate class logits. By balancing detailed spatial information with global semantic understanding, the architecture is designed to improve segmentation accuracy compared to the linear probing approach. Training utilizes binary cross-entropy with logits loss.

## 1.3. Linear Probing for Classification and Regression

The linear probing classifier consists of a single linear layer that takes high-dimensional CLS embeddings and projects them to the binary classes. The model uses binary cross-entropy with logits loss for binary classification tasks and mean squared error (MSE) loss for regression. It's designed to be used for evaluating the quality of pre-trained CLS embeddings by measuring how well they can be linearly separated for downstream tasks.

**Table S1.** The results of pairwise comparisons on different tasks between the foundation models.

| Task | Scenario | Metric | Model0 | Model1 | p Value | Significant |
|---|---|---|---|---|---|---|
| **Segmentation (Linear Probing)** | Pneumothorax | Dice | CheXagent | RAD-DINO | 0.039 | FALSE |
| | | | CheXagent | BiomedCLIP | <0.0001 | TRUE |
| | | | RAD-DINO | BiomedCLIP | <0.0001 | TRUE |
| | | IoU | CheXagent | RAD-DINO | 0.0668 | FALSE |
| | | | CheXagent | BiomedCLIP | <0.0001 | TRUE |

| | | | Model 1 | Model 2 | p-value | Significant |
|---|---|---|---|---|---|---|
| | Heart | Dice | RAD-DINO | BiomedCLIP | <0.0001 | TRUE |
| | | | CheXagent | RAD-DINO | <0.0001 | TRUE |
| | | | CheXagent | BiomedCLIP | <0.0001 | TRUE |
| | | IoU | RAD-DINO | BiomedCLIP | <0.0001 | TRUE |
| | | | CheXagent | RAD-DINO | <0.0001 | TRUE |
| | | | CheXagent | BiomedCLIP | <0.0001 | TRUE |
| | | | RAD-DINO | BiomedCLIP | <0.0001 | TRUE |
| Segmentation (Custom Model) | Pneumothorax | Dice | CheXagent | RAD-DINO | 0.0014 | TRUE |
| | | | CheXagent | BiomedCLIP | <0.0001 | TRUE |
| | | | RAD-DINO | BiomedCLIP | <0.0001 | TRUE |
| | | IoU | CheXagent | RAD-DINO | 0.0056 | TRUE |
| | | | CheXagent | BiomedCLIP | <0.0001 | TRUE |
| | | | RAD-DINO | BiomedCLIP | <0.0001 | TRUE |
| | Heart | Dice | CheXagent | RAD-DINO | <0.0001 | TRUE |
| | | | CheXagent | BiomedCLIP | <0.0001 | TRUE |
| | | | RAD-DINO | BiomedCLIP | <0.0001 | TRUE |
| | | IoU | CheXagent | RAD-DINO | <0.0001 | TRUE |
| | | | CheXagent | BiomedCLIP | <0.0001 | TRUE |
| | | | RAD-DINO | BiomedCLIP | <0.0001 | TRUE |
| Classification | Pneumothorax | AUPRC | CheXagent | RAD-DINO | 0.0006 | TRUE |
| | | | CheXagent | BiomedCLIP | <0.0001 | TRUE |
| | | | RAD-DINO | BiomedCLIP | 0.027 | FALSE |
| | | AUROC | CheXagent | RAD-DINO | 0.0002 | TRUE |
| | | | CheXagent | BiomedCLIP | <0.0001 | TRUE |
| | | | RAD-DINO | BiomedCLIP | 0.0002 | TRUE |
| | Cardiomegaly | AUPRC | CheXagent | RAD-DINO | 0.0382 | FALSE |
| | | | CheXagent | BiomedCLIP | <0.0001 | TRUE |
| | | | RAD-DINO | BiomedCLIP | 0.0008 | TRUE |
| | | AUROC | CheXagent | RAD-DINO | 0.001 | TRUE |
| | | | CheXagent | BiomedCLIP | <0.0001 | TRUE |
| | | | RAD-DINO | BiomedCLIP | <0.0001 | TRUE |
| Regression | CTR | MSE | CheXagent | RAD-DINO | 0.3622 | FALSE |
| | | | CheXagent | BiomedCLIP | <0.0001 | TRUE |
| | | | RAD-DINO | BiomedCLIP | <0.0001 | TRUE |
| | | R2 | CheXagent | RAD-DINO | 0.4314 | FALSE |
| | | | CheXagent | BiomedCLIP | <0.0001 | TRUE |
| | | | RAD-DINO | BiomedCLIP | <0.0001 | TRUE |

**Supplementary Reference**